\documentclass[manuscript,screen]{acmart}

\AtBeginDocument{%
  \providecommand\BibTeX{{%
    \normalfont B\kern-0.5em{\scshape i\kern-0.25em b}\kern-0.8em\TeX}}}
\usepackage{graphicx}
\usepackage{subcaption}
\usepackage{array}
\usepackage{threeparttable}
\usepackage{longtable}

\setcopyright{acmcopyright}
\copyrightyear{2018}
\acmYear{2023}
\acmDOI{XXXXXXX.XXXXXXX}

\acmPrice{15.00}
\acmISBN{978-1-4503-XXXX-X/18/06}

\begin{document}

\title[A Review of Digital Learning Environments for Teaching NLP in K-12]{A Review of Digital Learning Environments for Teaching Natural Language Processing in K-12 Education}

\author{Xiaoyi Tian}
\orcid{0000-0002-5045-0136}
\email{tianx@ufl.edu}

\affiliation{%
  \institution{University of Florida}
  % \streetaddress{P.O. Box 1212}
  % \city{Dublin}
  % \state{Ohio}
  \country{USA}
  % \postcode{43017-6221}
}
\author{Kristy Elizabeth Boyer}

\email{keboyer@ufl.edu}

\affiliation{%
  \institution{University of Florida}
  % \streetaddress{P.O. Box 1212}
  % \city{Dublin}
  % \state{Ohio}
  \country{USA}
  % \postcode{43017-6221}
}

\renewcommand{\shortauthors}{Xiaoyi Tian and Kristy Elizabeth Boyer}

\begin{abstract}
  Natural Language Processing (NLP) plays a significant role in our daily lives and has become an essential part of Artificial Intelligence (AI) education in K-12. As children grow up with NLP-powered applications, it is crucial to introduce NLP concepts to them, fostering their understanding of language processing, language generation, and ethical implications of AI and NLP. This paper presents a comprehensive review of digital learning environments for teaching NLP in K-12. Specifically, it explores existing digital learning tools, discusses how they support specific NLP tasks and procedures, and investigates their explainability and evaluation results in educational contexts. By examining the strengths and limitations of these tools, this literature review sheds light on the current state of NLP learning tools in K-12 education. It aims to guide future research efforts to refine existing tools, develop new ones, and explore more effective and inclusive strategies for integrating NLP into K-12 educational contexts. 
\end{abstract}


\maketitle

\section{Introduction}
Natural Language Processing (NLP) is integrated into our everyday lives ubiquitously, with applications including spell checkers, machine translation, recommendation systems and conversational agents. To ensure that citizens, including young people, become responsible users and creators of intelligent solutions, there has been increasing attention towards incorporating artificial intelligence (AI) curricula, including NLP, into 21st century computing education \cite{eguchi2021contextualizing,touretzky2019envisioning,wong2020broadening}. Children are growing up with NLP-powered applications, making them ideal learning platforms. For example, conversational agents have been demonstrated to enhance engagement in reading \cite{xu2020exploring}, foster language learning \cite{xu2021current,gomez2013video} and promote story comprehension and engagement \cite{xu2022dialogue}. 

Despite the prevalence of NLP in young people’s lives, there is limited research on teaching them NLP concepts. Traditionally, AI and NLP concepts have been taught primarily in higher education \cite{alm2021visualizing,litman2016natural,register2020learning}. However, research indicates that children are capable of grasping AI and NLP concepts from a young age \cite{hitron2019can,druga2021children}. Exposure to such knowledge is crucial for young people to understand how language is processed and generated, how and where NLP systems can be applied, and the social, economical and ethical challenges arising from AI and NLP use \cite{lin2020zhorai}. Fostering AI literacy can inspire more students to consider careers in AI and NLP, laying a strong foundation for their higher education and professional lives \cite{lee2021developing}.

There have been efforts to establish frameworks for AI and NLP education at the K-12 level. \citet{touretzky2019envisioning} founded the AI4K12 initiative, which proposed five K-12 AI guidelines for what AI concepts should every child know: 1) the perception and interpretation of complex information by AI systems, 2) their representation and reasoning capabilities, 3) the ability of AI systems to learn and improve, 4) their capacity to make natural interaction with humans, and 5) the broader societal implications and ethical considerations of AI. These ``big ideas'' provide a comprehensive yet succinct roadmap to navigate the landscape of AI education.

In this landscape, NLP emerges as a crucial element in AI education due to its role in facilitating machines' understanding, interpretation and generation of human language \cite{touretzky2022artificial,NaturalLanguageProcesing4All}. The AI4K12 big ideas highlight many NLP tasks and applications for children to grasp. According to \citet{touretzky2019envisioning}, students should understand basic NLP concepts such as speech recognition (big idea \#1), word embeddings (big idea \#2), parsing (big idea \#3), text generation and sentiment analysis (big idea \#4), as well as ethical concerns related to NLP applications (big idea \#5). To transform students from AI consumers into AI creators, it is essential for learning environments to enable students to have authentic, hands-on learning experiences \cite{shaffer1999thick, long2020ai}. Such experiences can be facilitated through relatable NLP tasks that simulate real-world applications, such as creating personalized chatbots and exploring sentiment analysis models.  

Learning NLP through developing real-world NLP-powered applications is challenging for novice learners, as many NLP tasks involve complex algorithms and require a high level of programming expertise. Common NLP tasks include sentiment analysis, text classification, topic modeling, machine translation, speech recognition, text-to-speech, part-of-speech tagging, and question answering \cite{jurafsky2019speech}. Typically, these tasks are executed using text-based programming languages (e.g., Python) and on platforms that require substantial computing resources (e.g., Tensorflow). Text-based programming interfaces and resource intensity might not align with the developmental capabilities of children or the resources of learning contexts, posing potential barriers to accessibility and comprehension \cite{garg2022last}. Thus, there is a critical need for intuitive and age-appropriate tools for introducing NLP concepts to younger learners.

Some digital learning environments aim to democratize AI by allowing novice users to create programs without extensive programming skills or substantial computing resources \cite{zimmermann2019youth}. These learning environments, such as Teachable Machine \cite{carney2020teachable}, AISpace2 \cite{zhou2020aispace2} and SmileyCluster \cite{wan2020smileycluster} help reduce users’ cognitive load and allow users to focus on the core AI concepts without having to deal with syntax issues or external libraries, making them useful tools for K-12 learners.

Inspired by the success of these digital learning environments in teaching general AI concepts, similar platforms have been developed for teaching NLP. These specialized tools typically support at least one NLP task, such as text classification or sentiment analysis, providing an intuitive introduction to the field. They may encompass a series of pipeline components, including data collection and preprocessing, model building, performance evaluation, and model exporting and deployment. Importantly, the capability for model deployment connects user-created NLP models to real-world applications, infusing a sense of relevance into K-12 computing education and enhancing student engagement \cite{wong2020broadening}.

To date, no literature review focuses on digital learning environments for teaching NLP in K-12 education. The most relevant review, presented by \citet{gresse2021visual}, examines tools for teaching machine learning in K-12. However, their work concentrates on tools that support development of ML models in general, with the majority centering image recognition models. It does not address NLP-specific tasks such as dialogue systems and speech synthesis. Furthermore, their review provides details on the tools' development, but the report on the empirical findings is relatively superficial in multiple aspects, including lack of information on learner backgrounds, assessments, and study outcomes. Other reviews such as \citet{lockwood2017computational} and \citet{taslibeyaz2020develop} are centered around tools and trends in teaching computational thinking, rather than AI or NLP. \citet{valko2022cloud} review AI teaching tools without specifically targeting the K-12 context. Reviews on teaching AI in K-12, such as \citet{lee2020analysis} and \citet{garcia2020learningml}, examine features of several learning tools. Meanwhile, \citet{giannakos2020games} provides a comprehensive summary of games designed for teaching machine learning, and \citet{zhou2020designing} highlights 49 AI-education works based on the AI literacy framework proposed by \citet{long2020ai}. Given the rapid advancement of NLP applications and the current gap in literature, this review is poised to provide a comprehensive examination of the learning tools that are designed for teaching NLP in K-12.

In this paper, we present a review of existing digital learning environments for teaching NLP in K-12. The purpose of this review is to explore available digital learning tools for teaching NLP and to understand how these environments support specific NLP tasks, NLP procedures, and their explainability. We will also describe how the tools and corresponding pedagogical activities are evaluated and provide a case study of a popular learning environment. The the contributions of this review can help educators and practitioners choose the most appropriate tool for their teaching needs and guide researchers in developing more effective and accessible NLP learning tools and activities. This review aims to answer the following research questions:
\begin{enumerate}

    \item What digital learning environments are available for NLP learning in K-12 education?
    \item What NLP learning tasks do these tools support, and how do they support them? 
    \item How have researchers evaluated these tools in educational contexts?
\end{enumerate}

In the remainder of this article, we will first describe the literature search procedure (section \ref{sec:lit-search}) which identifies 11 learning environments across 21 articles. Next, in section \ref{sec:results}, we will present the results, organized according to each research question. This will entail a detailed description of the systems' characteristics and their respective implementation strategies, followed by findings from the evaluation studies. To demonstrate how a particular system is developed and evaluated, section \ref{sec:case-study} will present a case study that offers practical insights. Lastly, in section \ref{sec:discussion}, we will pinpoint research gaps derived from the outcomes of this literature review and discuss their implications for K-12 NLP education. 

\section{Literature search} \label{sec:lit-search}
The objective of this study is to investigate the state of the art in digital learning environments for learning NLP in the context of K-12. We aim to characterize and compare the implementation and evaluation of these tools to identify gaps and potential opportunities for future research. 

The literature review was conducted using a non-systematic approach to search and review existing literature. Based on the research questions, we identified two main term-categories to include in the literature search: discipline (NLP-related) and target population (K-12). After performing multiple iterations of searches, we derived a list of relevant synonyms for each category. The search terms were applied over both the titles as well as the abstracts of the publications. In the literature search, we used the combined keywords from the two categories (Table \ref{table:keyword-list}). The complete search string was (``AI Learning'' AND ``AI Education'' AND ``AI literacy'' AND ``NLP'' AND ``natural language processing'' AND ``linguistics'' AND ``conversational AI'' AND ``dialog* system'' AND ``chatbot'') OR (``K12'' AND ``middle school'' AND ``high school'' AND ``elementary'' AND ``primary school'' AND ``secondary education'' AND ``youth'' AND ``kid'' AND ``child*''). The actual string varied based on the restrictions of each database.  
\begin{table}[h]

\caption{Keyword list for literature search}
\label{table:keyword-list}
\begin{tabular}{|p{0.2\linewidth}|p{0.7\linewidth}|}
\hline
\textbf{Category (connected using ``and'' logic)} & \textbf{Search terms (connected using ``or'' logic)   }                                                                                        \\ \hline
Discipline                             & AI Learning, AI Education, AI literacy, NLP/natural language processing, linguistics, conversational AI, dialog(ue) system, chatbot \\ \hline
Target population                      & K12, middle school, high school, elementary/primary school, secondary education, youth, kid, child*                                 \\ \hline
\end{tabular}

\end{table}

\subsection{Sources} 

We searched the main digital databases and libraries in the field of computing, including ACM Digital Library, IEEE Xplore, ScienceDirect and Google Scholar. In addition, we also used Google search to account for the possibility that some educational tools may not yet have been published in scientific databases \cite{piasecki2018google}. Since research on the topic of NLP education is fairly new, recent works are most often published in niche conferences and workshops, including AAAI Symposium on Educational Advances in Artificial Intelligence (EAAI), Special Interest Group on Computer Science Education Technical Symposium (SIGCSE), Interaction Design and Children (IDC), ACM CHI Conference on Human Factors in Computing Systems (CHI), Computer-Supported Cooperative Work (CSCW), IEEE Symposium on Visual Languages and Human-Centric Computing (VL/HCC). We performed additional searches in these proceedings to minimize the risk of omitting relevant works.

\subsection{Selection criteria}

Based on the goal of this literature review, following the criteria described in \citet{tatar2019literature}, the selection criteria for this literature review are shown in Table \ref{table:selection-criteria}. 

\begin{figure}[!htb]
  \includegraphics[width=0.6\textwidth]{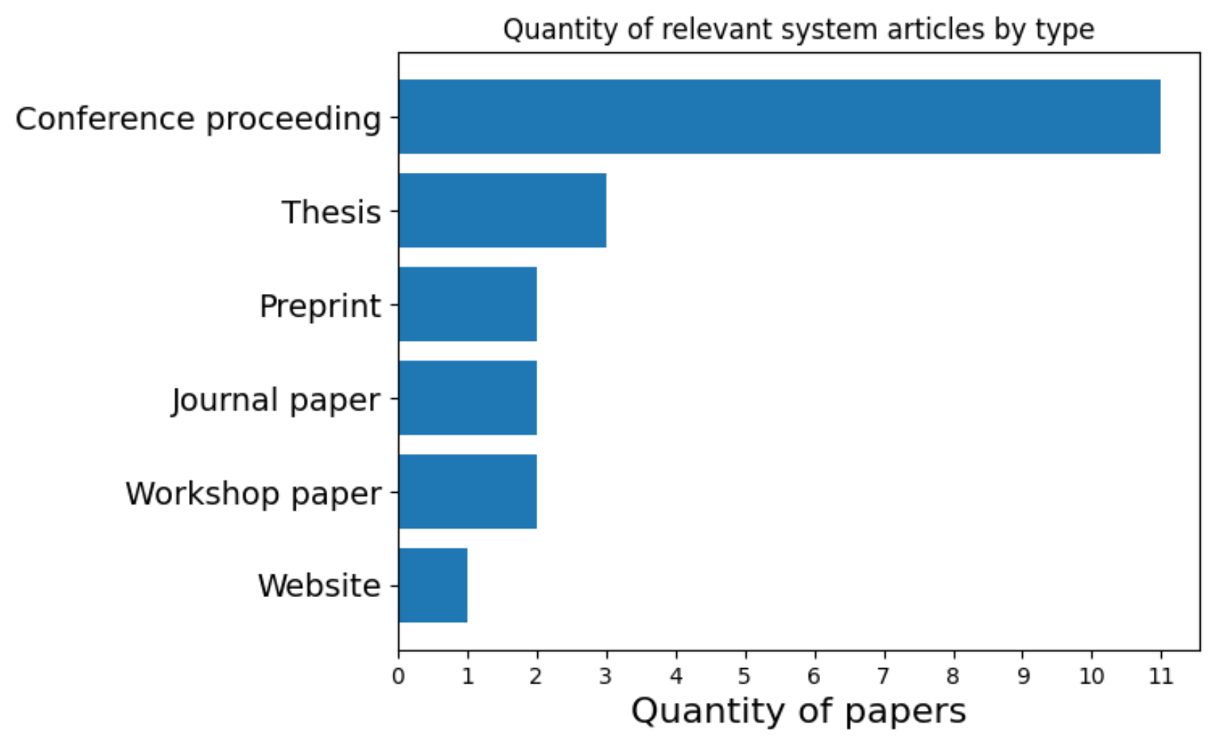}
  \caption{Distribution of types of the relevant reviewed publications}

  \label{fig:paper_type}
\end{figure}

\begin{table}[h]
\caption{Selection criteria for NLP learning tools for K-12}
\label{table:selection-criteria}

\begin{tabular}{|p{0.4\linewidth}|p{0.4\linewidth}|}
\hline
\textbf{Inclusion criteria}                                                               & \textbf{Exclusion criteria}                                                                                     \\ \hline
K12 education (kindergarten through the 12th grade)                                       & Other stages of education such as pre-university level, college, and graduate level                             \\ \hline
Empirical studies                                                                         & Theoretical studies                                                                                             \\ \hline
Involves tools or technologies                                                            & Studies that do not involve digital tools (e.g., curriculum design, unplugged activities only, workshop design) \\ \hline
Report at least one form of assessment (e.g., learning outcomes, engagement, perception)* & Studies that do not provide assessment*                                                                         \\ \hline
English publications                                                                      & Non-English publications                                                                                        \\ \hline
\end{tabular}

\begin{tablenotes}
\item \small
    Note. Criteria marked with an asterisk (*) were used for selection of papers answering RQ3 only 
\end{tablenotes}

\end{table}
The initial literature search was conducted in March 2022. A second search was conducted in February 2023 to include any additional papers published between March 2022 and February 2023. The searched papers were screened by scrutinizing their titles and abstracts to determine their eligibility based on the selection criteria. Because this field is still new, some learning environments are still works-in-progress and thus lack a published system evaluation. However, it is still important to include these tools for the completeness of this review. For tools that only include system implementation but not user evaluation, we have considered them to answer RQ1 and RQ2, but not RQ3. Through backwards and forwards snowball sampling, this review finally yielded 21 publications describing 11 learning environments. Their formats varied widely, including websites, theses, and conference proceedings. The distribution of publications by type is shown in Figure \ref{fig:paper_type}.

\section{Results} \label{sec:results}

\subsection{What digital learning environments are available for NLP learning in K-12 education?}

We identified 11 digital learning environments developed for NLP learning in K-12 education. These learning environments and corresponding evaluation studies are found in 21 publications (some systems involve multiple studies published as different papers). Below we will briefly introduce each learning tool, its background, functionalities, and the learning topics covered by the tool. 

\begin{enumerate}

\item \textbf{NLP4ALL \cite{baglini-hjorth-2021-natural, NaturalLanguageProcesing4All}:} Developed at Aarhus University, Denmark, NLP4All is a natural language processing tool that facilitates the analysis of tweets from political parties, helping students with no prior ML experience to explore differences and similarities between policy views and communication styles.

\item \textbf{Cognimates \cite{druga2018growing}:} Initially created by a student at the MIT Media Lab, Cognimates is intended for children aged 6-14. This platform introduces machine learning and image/text classification concepts. The block programming platform "Codelab" further expands learning opportunities with block sections about feelings, Twitter, text training, image training, and translation.

\item \textbf{eCraft2Learn Unified User Interface \cite{kahn2018a,kahn2018ai,alimisis2018stem,alimisis2019kids}}: This interface, including the Snap! blocks, is part of the broad eCraft2Learn project, funded by the European Union and led by the University of Eastern Finland. This platform offers introduction to robotics, DIY electronics, visual programming, 3D modelling, and 3D printing to learners above age of 7. It facilitates hands-on learning experiences and deepens understanding of technology and fabrication. Besides the eCraft2Learn interface, this project also provides a comprehensive teacher guide, student worksheet, activities and information about the hardware on the project website\footnote{https://project.ecraft2learn.eu/}. 

\item \textbf{Machine Learning for Kids (ML4Kids) \cite{lane_2018}:} Dale Lane, a developer at IBM, initially developed this tool as a personal project. ML4Kids introduces AI and ML concepts to children at all ages. It offers a learning environment where children can learn the basics of ML and develop their own machine learning models for recognizing text, images, numbers, and sounds. The site also contains a collage of tutorials of machine learning concepts (e.g., How to use confidence scores), step-by-step guides of different activities, as well as visualizations of user-created models that scaffold the deep learning processes. 

\item \textbf{Teachable Machine (TM) \cite{carney2020teachable,toivonen2020co}:} Google's Creative Lab created Teachable Machine as a web-based interface for users of all ages. It empowers users to train their own image, audio, or pose-based ML classification models without coding. The tool offers a practical introduction to machine learning, image/pose/audio classification, and deep learning.

\item \textbf{LearningML \cite{garcia2020learningml,rodriguez2021evaluation}:} LearningML is developed by a group of educational technologists in Spain. It is a platform that simplifies the process of building supervised machine learning models and learning about ML classification. It creates an interactive and educational environment that encourages students between 10-16 years old to explore machine learning.

\item \textbf{Convo \cite{zhu2021teaching,zhu2021creating}:} Two graduate students at MIT developed Convo for middle school students. As a conversational programming agent, Convo enables students to create deep learning-based conversational AI agents. It provides a learning environment that explores AI-driven communication systems and their applications.

\item \textbf{Zhorai \cite{lin2020zhorai}:} Created by researchers at both Harvard Graduate School of Design and MIT,  Zhorai is a conversational agent that teaches AI/ML concepts through interactive dialogue for young users (aged 8-11). It focuses on representation and reasoning, learning, and the social impact of AI.

\item \textbf{ConvoBlocks \cite{van2021teaching,van2019tools,van2021alexa,van2022learning}:} ConvoBlocks is a block-based programming interface developed by MIT for learners between the ages of 11-18. It offers a hands-on experience in training, transfer learning, large language models, intents, societal impact and ethics, speech synthesis, and speech recognition.

\item \textbf{Interactive Word Embeddings (IWE) \cite{bandyopadhyay2022interactive}:} IWE was designed in a collaborative effort of University of Maryland and Carnegie Mellon University for high school and college students. This tool allows  students to build their understanding of preprocessing in NLP by visually exploring word embeddings, relationships, analogies, and semantics in vector space. 

\item \textbf{Build-a-Bot \cite{pearce2022build}:} Developed by researchers at MIT, Build-a-Bot is an open-source tool designed for classroom environments. It introduces students to the NLP pipeline, which includes data collection and labeling, data augmentation, keyword filtering, intent recognition, and question answering, serving as a valuable resource for teaching AI concepts.

\end{enumerate}

\begin{figure}[!htb]
  \includegraphics[width=0.95\textwidth]{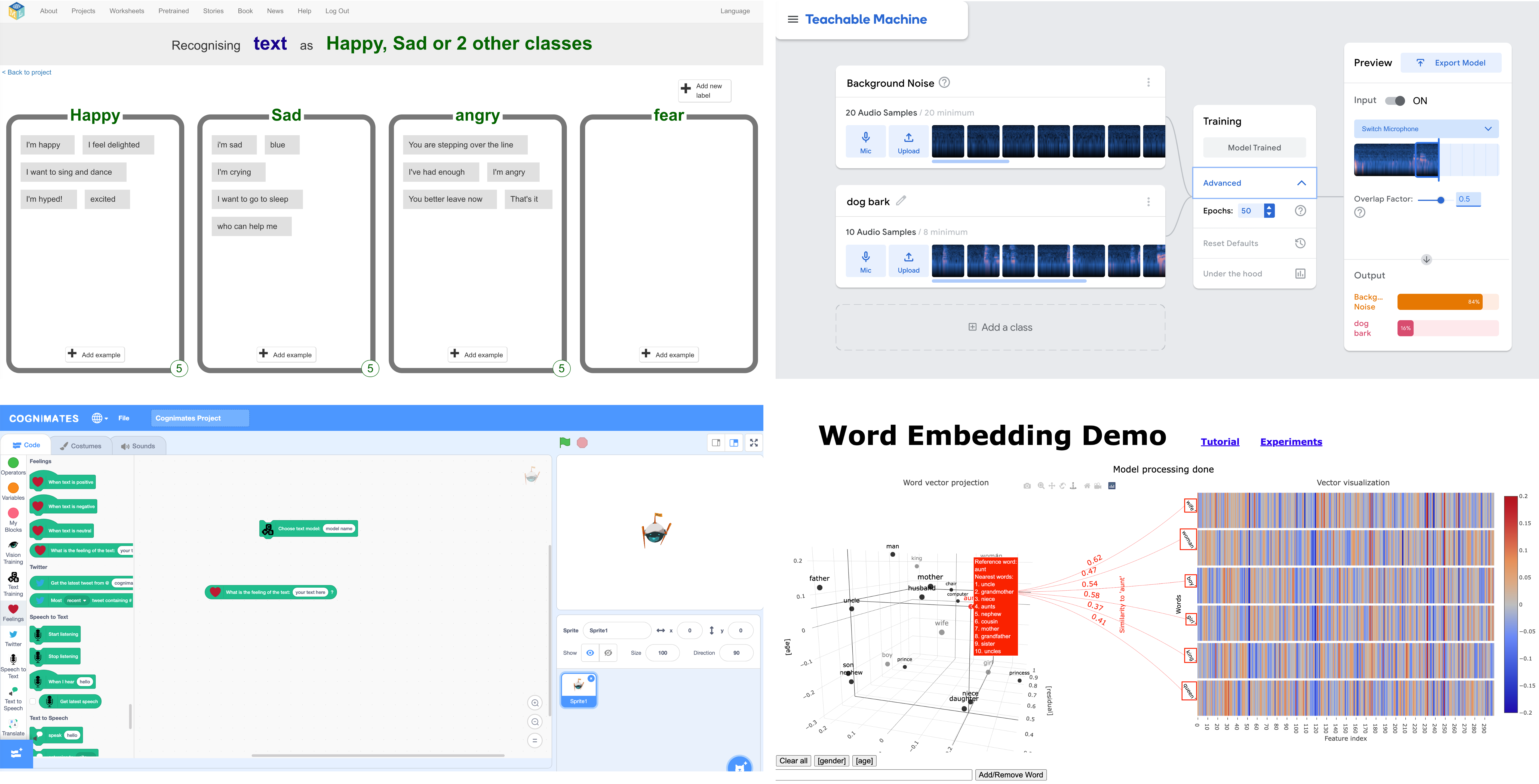}
  \caption{Screenshots of example learning environments supporting NLP tasks. Top left: Machine Learning for Kids; Top right: Teachable Machine; Bottom left: Cognimates; Bottom right: Interactive Word Embeddings}

  \label{fig:tool-screenshots}
\end{figure}

A summary of these learning environments is presented in Table \ref{table:basic-infor} and screenshots of some exemplar learning systems (ML4Kids, TM, Cognimates, IWE) are shown in Figure \ref{fig:tool-screenshots} (top left, top right, bottom left, and bottom right, respectively). Most of these tools are available as web applications, which makes them easily accessible to anyone with an internet connection. However, some tools (e.g., Convo) are not publicly available, which limits their accessibility. Of the above tools that are publicly available, most are available for free, but some are restricted to registered users (e.g., NLP4All, eCraft2Learn) or require an API key to access them (e.g., Cognimates). 7 out of 11 tools allow users to deploy the artifact (a working application or a trained model) to an external site. Three tools do not offer external integration and one tool does not mention integration. Regarding language support, six tools only support English, while five support at least one more language in addition to English. Among those five tools that support more than one language, three tools offer multiple (10+) choices.

\begin{table}[h]
  \caption{Basic information for all tools}

  {

\begin{tabular}{|>
{\raggedright\arraybackslash}p{1.6cm}|>{\raggedright\arraybackslash}p{6cm}|>{\raggedright\arraybackslash}p{3cm}|>{\raggedright\arraybackslash}p{2cm}|}
\hline
\textbf{Name} & \textbf{Website} & \textbf{Integration} & \textbf{Language} \\
\hline
NLP4ALL & \url{http://86.52.121.12:5000/} &  no external integration & English, Danish \\
\hline
Cognimates & \url{cognimates.me/home} &  Codelab (a scratch fork) & Multiple languages \\
\hline
eCraft2Learn & \url{https://ecraft2learn.github.io/uui/index.html} &  Snap! & English, Indonesian \\
\hline
ML4Kids & \url{https://machinelearningforkids.co.uk/} & Scratch, App Inventor, Python & Multiple \\
\hline
TM & \url{https://teachablemachine.withgoogle.com/}  & TensorFlow.js or TensorFlow Lite & English \\
\hline
LearningML & \url{learningml.org} & Scratch; or export as Tensorflow.js & English, Spanish \\
\hline
Convo & not available &  - & English \\
\hline
Zhorai & \url{zhorai.csail.mit.edu} &  no integration & English \\
\hline
ConvoBlocks & \url{http://alexa.appinventor.mit.edu/} & Alexa-enabled devices & Multiple \\
\hline
IWE & \url{https://www.cs.cmu.edu/~dst/WordEmbeddingDemo/} & no integration & English \\
\hline
Build-a-Bot & not available  & PySimpleGUI, chatbotAI library & English \\
\hline
\end{tabular}

}
  \label{table:basic-infor}
\end{table}

\subsection{What NLP learning tasks do these tools support and how do they support them? }

In this section, we will explore the array of NLP tasks supported by these educational tools, considering their handling of various data types, input and output modalities, methods of NLP evaluation, and the level of explainability in these tools. A summary of the technical implementation of the tools is in Table \ref{table:tech-implement}.

\begin{table}[h]
  \caption{Technical implementation summary for all tools}

{
\centering
\begin{tabular}
{|>{\raggedright\arraybackslash}p{1.7cm}|>{\raggedright\arraybackslash}p{2.8cm}|>{\raggedright\arraybackslash}p{3.8cm}|>{\raggedright\arraybackslash}p{5.5cm}|}

\hline
\textbf{Tool} & \textbf{Supported NLP tasks} & \textbf{Backend} & \textbf{NLP evaluation metrics} \\ \hline
NLP4ALL & text classification & Naive Bayesian Classifier & classification confusion matrix \\ \hline
Cognimates & text classification, sentiment detection, dialogue & uClassify for text; clarifAI for image & classification result with confidence score based on user-input data \\ \hline
eCraft2Learn & speech synthesis, speech recognition & Pretrained cloud models, built-in browser support, IBM Watson & None \\ \hline
ML4Kids & text classification & IBM cloud & Classification result with confidence score based on user-input data \\ \hline
Teachable Machine & speech recognition & TensorFlow.js for building custom machine learning models & Training accuracy and loss function \\ \hline
LearningML & text classification & Tensorflow.js; uses Artificial Neural Network for ML; with additional transfer learning technique & Beginner mode: shows degree of confidence for one data entry using natural language. Advanced mode: shows the numerical probability of a label, expressed as a percentage, as well as a confusion matrix \\ \hline
Convo & intent recognition, entity recognition & Rasa & None \\ \hline
Zhorai & speech recognition, semantic parsing, text classification, speech synthesis & Web speech API; NLTK and Stanford CoreNLP toolkits; D3.js & Similarity ranking for classification results. Word visualization for NLU. \\ \hline
ConvoBlocks & dialogue, intent recognition, entity recognition & Amazon Web Services; App Inventor cloud & No specific evaluation metrics. Students can test the agent and see the intent recognition result. \\ \hline
Interactive Word Embeddings (IWE) & text preprocessing & Word2vec, with word vectors trained on a mix of Wikipedia text and news stories \cite{mikolov2018advances} & None \\ \hline
Build-a-Bot & question answering, intent recognition & BERT for intent recognition; DistilBERT and T5 for question answering & No evaluations. The tool itself provides a sample dataset about earth science for teachers and students to explore. \\ \hline
\end{tabular}

}

  \label{table:tech-implement}
\end{table}

\begin{figure}[!htb]
  \includegraphics[width=0.75\textwidth]{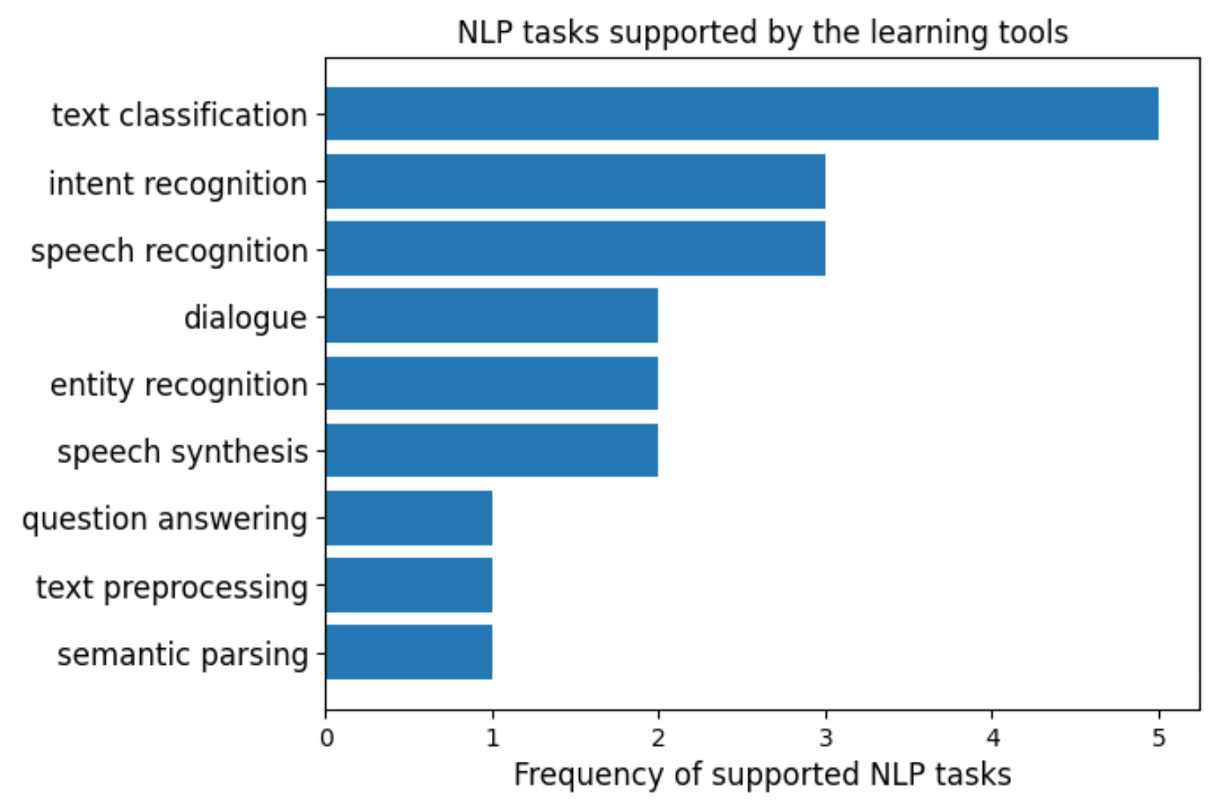}
  \caption{Frequency of supported NLP tasks. Many tools support more than one NLP task.}

  \label{fig:NLP_tasks}
\end{figure}

As shown in Figure \ref{fig:NLP_tasks}, text classification is the NLP task most commonly supported by these learning tools, with a total of 5 tools supporting it. Following text classification, speech recognition and intent recognition are supported by 3 tools each. Meanwhile, speech synthesis and entity recognition are supported by 2 tools, while semantic parsing, question answering, dialogue system, and text preprocessing are each supported by only 1 tool.

In the context of NLP education, illustrating the entire NLP process for learners is crucial, as it provides a comprehensive understanding of the inner workings of AI and how different components contribute to the overall outcome. Similar to the machine learning process, a typical NLP/ML system follows a four-step process: 1) generating training data, 2) building the model, 3) evaluating the model, and 4) exporting and deploying \cite{gresse2021visual}. Based on these general steps, we found that these 11 tools offer a range of capabilities for training and deploying NLP models. With the exception of one system (IWE \cite{bandyopadhyay2022interactive}), all the identified systems support users generating their own training data (step 1), allowing them either to collect and upload the text data by themselves or to manually label system-provided data. Only four systems (i.e., ConvoBlocks, LearningML, ML4Kids, Teachable Machine) support all four processes involved in building an NLP model. 

Cognimates \cite{druga2018growing} supports three of the generalized NLP steps, except for training (the interface was not specific about the process of training). Similarly, Zhorai \cite{lin2020zhorai} allows users to add their own training data, adjust training parameters, and explore existing models using mind map visualization. NLP4All \cite{baglini-hjorth-2021-natural} supports two modes: text tagging and robot training, allowing users to either manually tag individual tweets or utilize the classifier to classify tweets and experiment the results by specifying different word features. eCraft2Learn \cite{kahn2018a} uses Snap! \cite{kahn2018ai}, a block-based programming environment that is enhanced with AI modules. eCraft2Learn provides options for customizing speech output, including tone, pitch, volume, rate, and language. Other tools which focus on conversational AI (Convo \cite{zhu2021teaching}, Build-a-Bot \cite{pearce2022build}, and ConvoBlocks \cite{van2021teaching}) enable users to create intents by generating training phrases and entities; then, each of these three systems support some combination of connecting the intents with procedures, data augmentation and keyword filtering, and deploying the system to an external device such as Alexa. 

\begin{figure}[!htb]
  \includegraphics[width=0.6\textwidth]{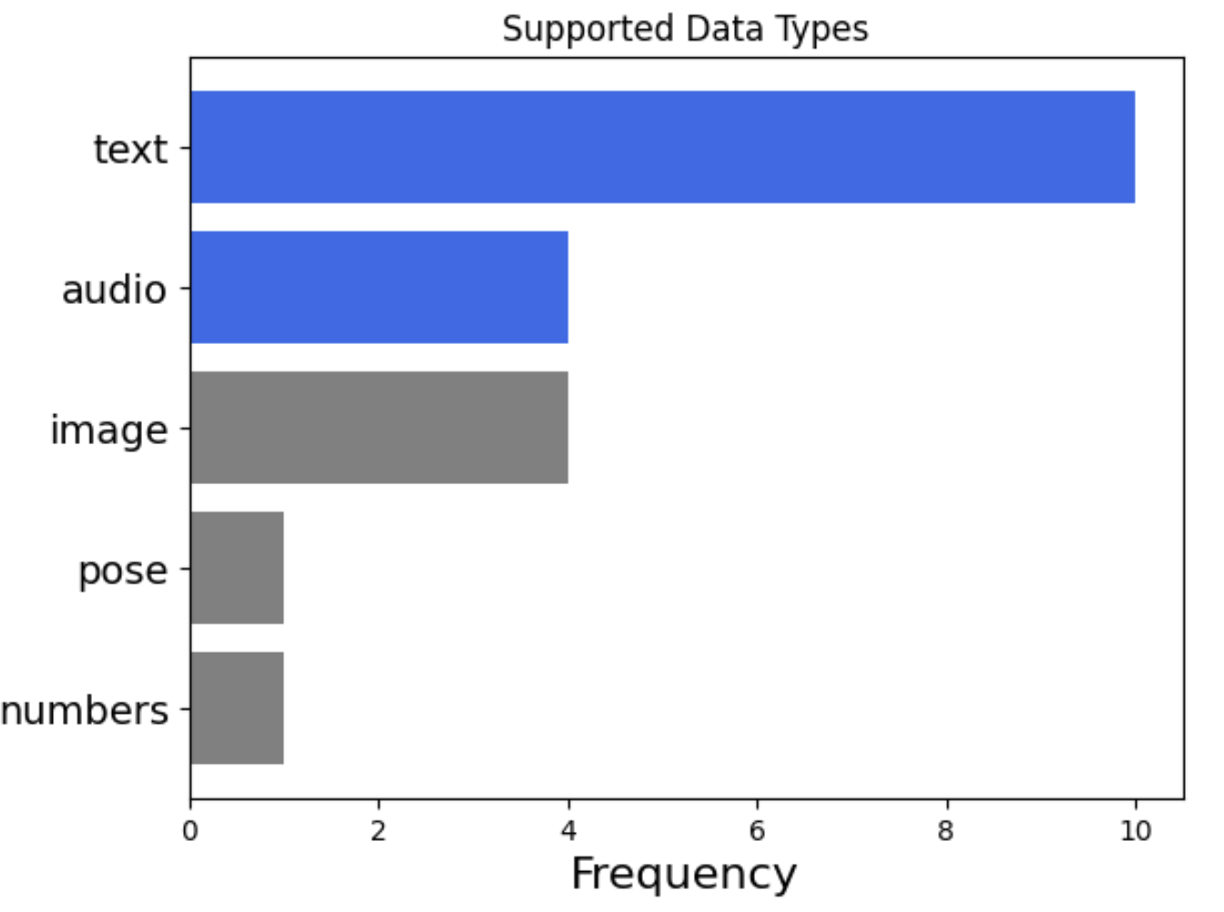}
  \caption{Frequency of supported data types. Some learning tools support multiple data types.}

  \label{fig:data_types}
\end{figure}

Next, to explore how these educational tools support different NLP tasks, we will discuss the different types of natural language data these tools accommodate, as well as the input and output modalities. Investigating the diversity of data modalities can help better understand the tools' adaptivity in different learning contexts and capacity for serving for diverse learners. In terms of data types (Figure \ref{fig:data_types}), textual data is the most common in NLP tasks: ten tools involve textual data and only four tools involve audio data. 

\begin{figure}[!htb]
  \includegraphics[width=0.7\textwidth]{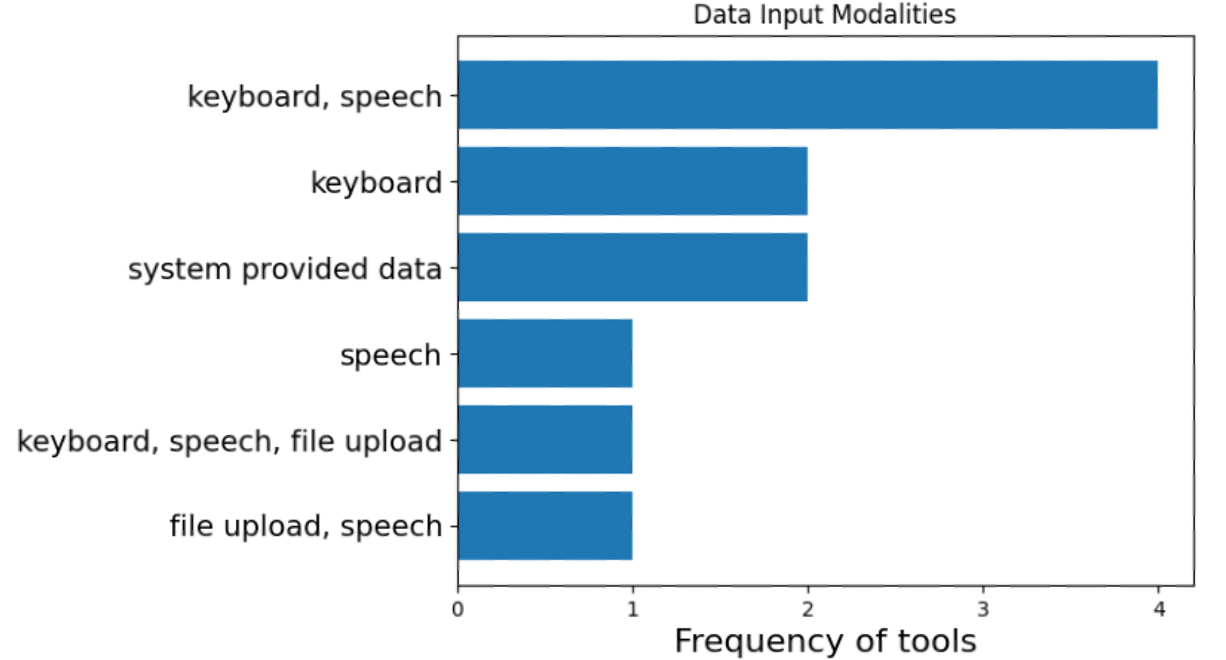}
  \caption{Data input modalities supported by the learning tools}

  \label{fig:input_mod}
\end{figure}

As for the data input modalities (Figure \ref{fig:input_mod}), text via keyboard is the most popular modality, with eight systems involving this modality. However, speech also appears to be a popular data input modality, with a total of seven systems involving it. This may reflect a design choice to tailor these systems to young users’ developmental level and abilities. Younger children may have limited typing skills and attention spans and may prefer more interactive input modalities such as speech. Allowing children to input data using their speech without typing can lower their frustration and be more engaging. There are two systems that provide existing data for the users to tinker with \cite{baglini-hjorth-2021-natural,bandyopadhyay2022interactive}. NLP4All provides 4500 tweets for students to explore the differences and similarities between different political parties; IWE contains word vectors trained on existing large datasets so that young users can explore the semantic relationships between different words in a 3-D space.  

\begin{figure}[!htb]
  \includegraphics[width=0.8\textwidth]{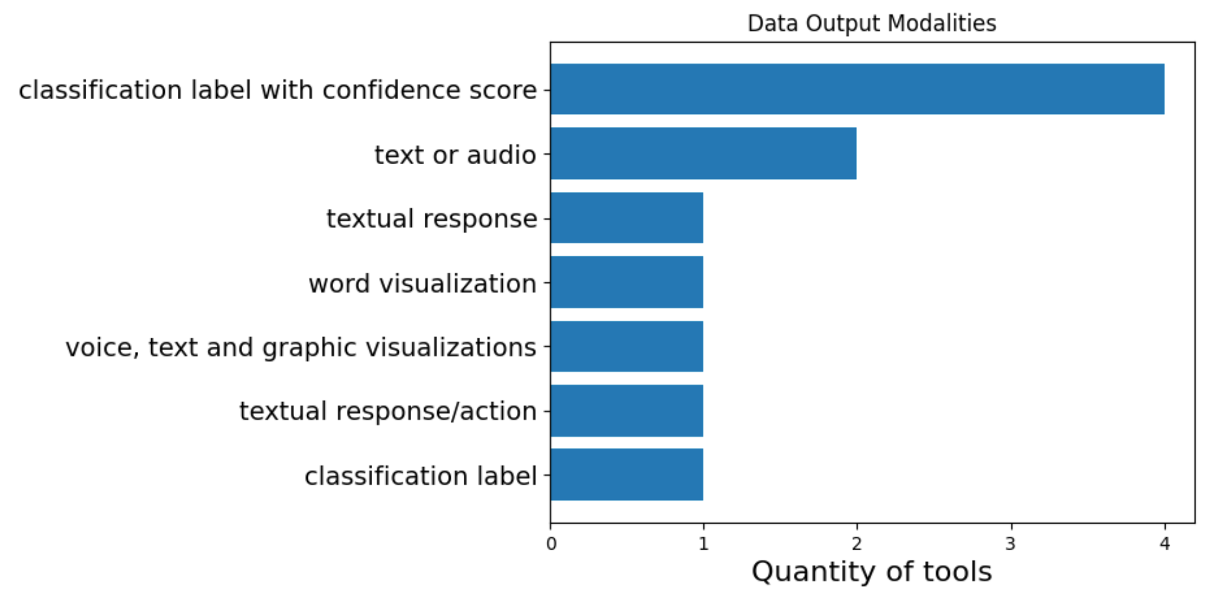}
  \caption{Data output modalities supported by the learning tools}

  \label{fig:output_mod}
\end{figure}

Figure \ref{fig:output_mod} shows the output modalities offered by the systems. Because the primary NLP task these tools support is text classification, the results of text classification are the most common data output modality. Among the five systems which provide classification results, four also indicate some kind of confidence score for the prediction, showing how confident the system is that it has correctly classified the input \cite{hagen2015twitter}. Because this is an important concept in machine learning, a few systems offer adaptivity or additional scaffolding regarding the \textit{confidence level} concept. For example, ML4Kids offers a series of tutorials explaining the ML concepts, including what is a confidence score and how to use it effectively. LearningML offers adaptivity for scaffolding this concept. For beginners, the system indicates various degrees of the confidence level in natural language such as ``not sure'' or ``probably''; if the user is in the advanced mode, the system will indicate the confidence score as numerical probability. One system (Build-a-bot) output text-only response and the remaining systems (eCraft2Learn, Zhorai, ConvoBlocks, IWE, Convo) include additional output modalities such as voice, action and graphic visualizations. 

\begin{figure}[!htb]
    \includegraphics[width=0.7\textwidth]{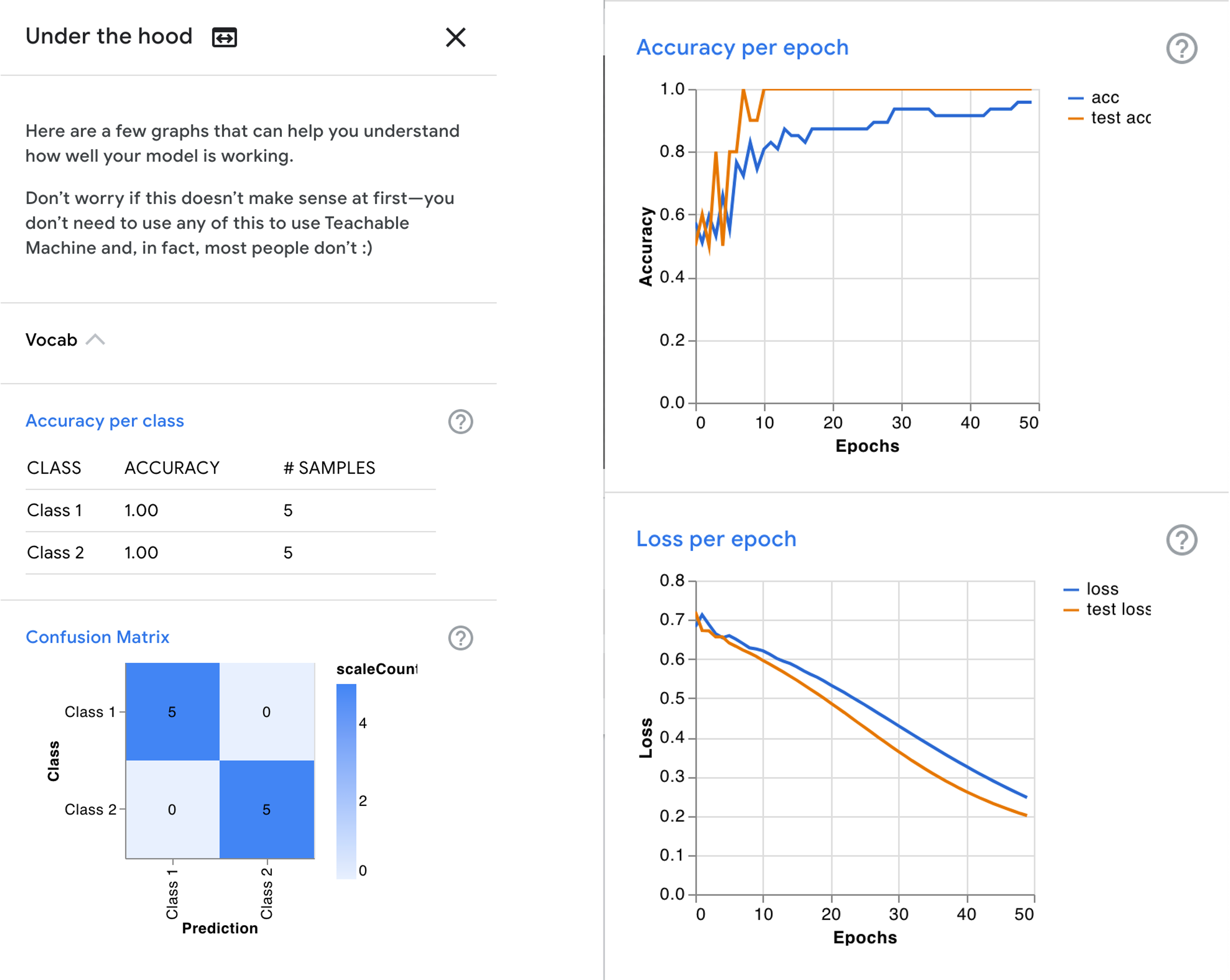}
    \caption{The ``under the hood'' pages in Teachable Machine explaining the accuracy and loss per learning epoch}
    \label{fig:tm-explainability}
\end{figure}

\begin{figure}[!htb]
    \includegraphics[width=0.85\textwidth]{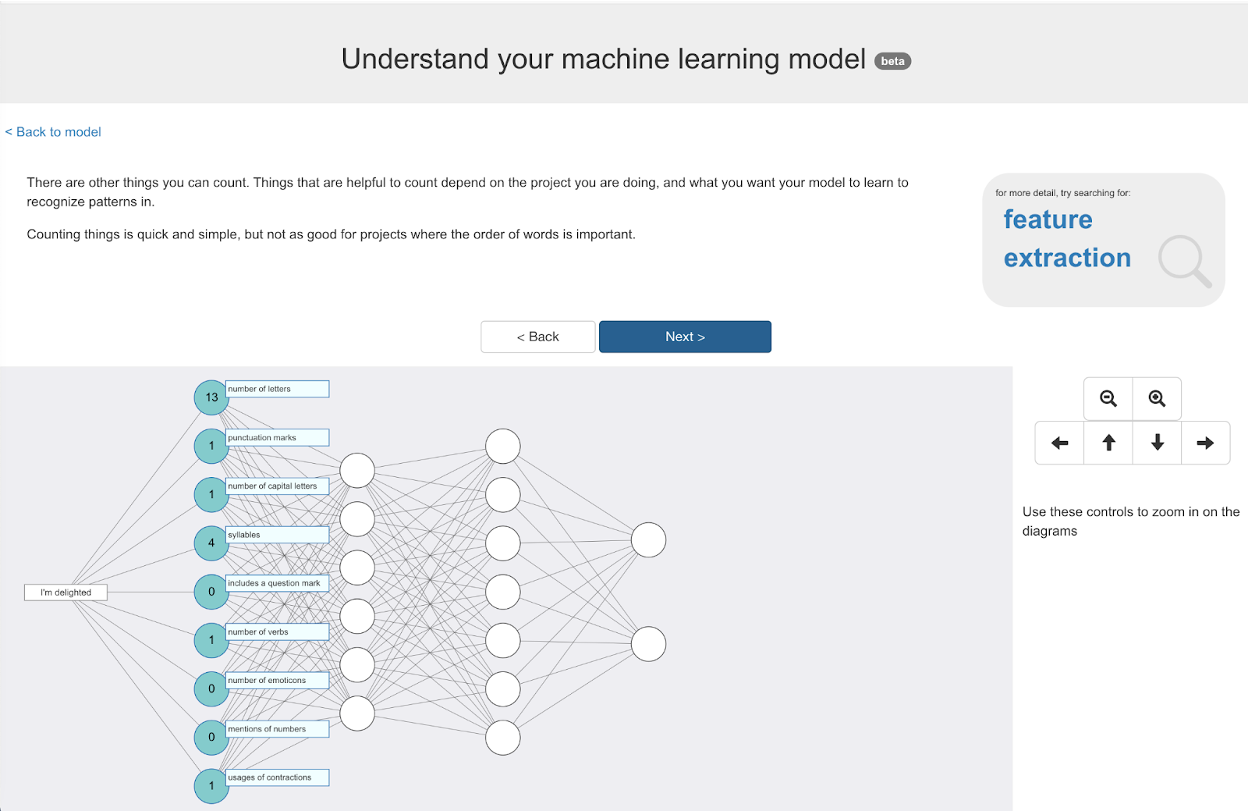}
    \caption{The ``describe your model'' module in ML4Kids that provides animated scaffolding of neural network concepts }
    \label{fig:ml4kids-explain}
\end{figure}

One important task for pedagogical systems is scaffolding what happens ``under-the-hood''. This involves helping learners to understand the NLP tasks and processes, explaining how the model works, why it works or does not work, and what leads the system or model to perform a certain way. Of the tools we reviewed, some systems provide ``output explanation'', often achieved by showing the evaluation metric along with some explanation, like confidence level. The most popular evaluation metric provided by these systems is model accuracy. Two (LearningNL, NLP4All) systems display a confusion matrix. Six systems do not provide specific evaluation or ``output explanation'' for their NLP tasks.  Some systems offer ``model explanation'', which allows the users to tinker the training models by specifying training features (e.g., NLP4All), customizing learning parameters (e.g., epochs in TM). Some systems provide additional scaffolding on the NLP process itself. For instance, TM has an option for the users to explore what’s happening ``under the hood'' by showing accuracy and loss per learning epoch (Figure \ref{fig:tm-explainability}). Similarly, in ML4Kids, there is a ``describe your model'' module that shows the animated scaffolding of neural network concepts including input layer, encoding, bag of words, feature extraction, word embedding, hidden layer, biases, weight, loss function, back propagation (Figure \ref{fig:ml4kids-explain}). ML4Kids also encourages learners to input more training data (``The more you can get, the better it should learn, but you need at least five examples of each as an absolute minimum.'').

% The scaffolding is still sort of text-heavy, which might not be attractive to younger learners. 

\subsection{How have researchers evaluated these tools in educational contexts? }

In this subsection, we will summarize the evaluation results of the learning systems (Table \ref{table:eval-study}). Among the 21 publications collected for this review, a total of fifteen reported on learning system evaluations, so this subsection is limited to the discussion of this subset. We will begin with a summary of the learners' characteristics, then move on to the types of research design and study contexts, and will finally cover the types of assessments and findings of the studies.

{

\begin{longtable}{|>{\raggedright\arraybackslash}p{1.6cm}|>{\raggedright\arraybackslash}p{0.5cm}|>{\raggedright\arraybackslash}p{3.2cm}|>{\raggedright\arraybackslash}p{1.2cm}|>{\raggedright\arraybackslash}p{2cm}|>{\raggedright\arraybackslash}p{2cm}|>{\raggedright\arraybackslash}p{2.5cm}|}
% {|p{1.6cm}|p{0.5cm}|p{3.2cm}|p{1.2cm}|p{2cm}|p{2cm}|p{2.5cm}|}

\caption{Evaluation studies of NLP learning tools} \label{table:eval-study}\\

\hline
\textbf{Tool} & \textbf{Ref.} & \textbf{Outcomes} & \textbf{Research design} & \textbf{Study location} & \textbf{Sample size and participant age} & \textbf{Application context} \\ 
\hline
\endfirsthead

\multicolumn{7}{c}%
{{\bfseries \tablename\ \thetable{} -- continued from previous page}} \\
\hline
\textbf{Tool} & \textbf{Ref.} & \textbf{Assessment type} & \textbf{Research design} & \textbf{Study location} & \textbf{Sample size and participant age} & \textbf{Application context} \\ \hline
\endhead

\hline \multicolumn{7}{|r|}{{Continued on next page}} \\ \hline
\endfoot

\hline
\endlastfoot

NLP4ALL & \cite{NaturalLanguageProcesing4All} & learning experience & Qual & Denmark & 24 high school students & formal (social studies classroom) \\ \cline{2-7} 
 & \cite{baglini-hjorth-2021-natural} & learning experience & Case study & Denmark & 20 masters students & formal, introductory computational linguistics class \\ \hline
Cognimates & \cite{druga2018growing} & AI knowledge, Perception of the agents & Mixed & Multiple (US, German, Denmark, Sweden) & 104, age 6-14 & informal \\ \hline
eCraft2Learn & \cite{kahn2018a} & learning experience & Case study & Sri Lanka and Singapore & 25 undergraduate, 18 children age 7-13 & informal (afterschool programs) \\ \cline{2-7} 
 & \cite{alimisis2019kids} & learning experience & Qual & Greece & 48, age 13-17 & informal \\ \cline{2-7} 
 & \cite{kahn2018ai} & AI knowledge (written essay); classroom experience & Mixed & Indonesia & 40, age 16-17 & formal classroom \\ \hline
ML4Kids & \cite{lane_2018} & no evaluation & - & - & - & - \\ \hline
Teachable Machine & \cite{toivonen2020co} & workshop experience & Qual & Finland  & 34 elementary students, age 12-13 & formal \\ \hline
LearningML & \cite{garcia2020learningml} & AI knowledge & Mixed & Spain (college) & 14 colleges students & informal (workshops) \\ \cline{2-7} 
 & \cite{rodriguez2021evaluation} & AI and ML knowledge (tested reliability and fidelity) & Mixed & Online & 135, age 10-16 & informal (online webinar and self-paced learning) \\ \hline
Convo & \cite{zhu2021teaching,zhu2021creating} & AI literacy, attitudes, perceptions about CA persona & Mixed & Online & 12, age 11-14 & informal \\ \hline
Zhorai & \cite{lin2020zhorai} & AI/ML knowledge; attitudes; interest/motivation in learning; engagement & Mixed & US & 14, age 8-11 & informal \\ \hline
ConvoBlocks & \cite{van2021teaching,van2019tools} & AI/ML knowledge/competency, student project topics; engagement and interest & Mixed & Online & 9 teachers and 47 students age 11-18 & informal \\ \cline{2-7} 
 & \cite{van2021alexa} & perceptions of Alexa persona; AI conception & Mixed & same as \cite{van2021teaching} & same as \cite{van2021teaching} & same as \cite{van2021teaching} \\ \cline{2-7} 
 & \cite{van2022learning} & perception and trust of CAs & Mixed & Online (US, Singapore, Canada, New Zealand, Indonesia, Iran, Japan, and India) & 27 children (age M = 13.96, SD = 1.829) and 19 parents (age M = 46.35, SD = 11.07) & informal, Zoom workshop \\ \hline
Interactive Word Embeddings (IWE) & \cite{bandyopadhyay2022interactive} & experience on the system & Case study & unknown & two high school students, one undergraduate & informal \\ \hline

Build-a-Bot & \cite{pearce2022build} & no evaluation & - & - & - & - \\ \hline

\end{longtable}
}

\subsubsection{Learners}
The majority of learning activities target middle and high school learners, who are approximately 11-18 years old. Only three studies involve children under the age of 11, which is roughly equivalent to 6th grade in the United States education system. Some studies also included adult learners, such as parents and undergraduate students, as part of their pilot study. Learners in the studies come from a variety of locations, including Denmark, the United States, Germany, Sweden, Sri Lanka, Singapore, Greece, Indonesia, Finland, Spain, and Canada. Some studies were conducted online with participants mainly from the US, and also from countries like Singapore, Canada, New Zealand, Indonesia, Iran, Japan, and India.

\subsubsection{Evaluation Methodologies and Study Contexts}

\begin{figure}[htbp]
  \centering
  \begin{minipage}[b]{0.45\textwidth}
    \includegraphics[width=\textwidth]{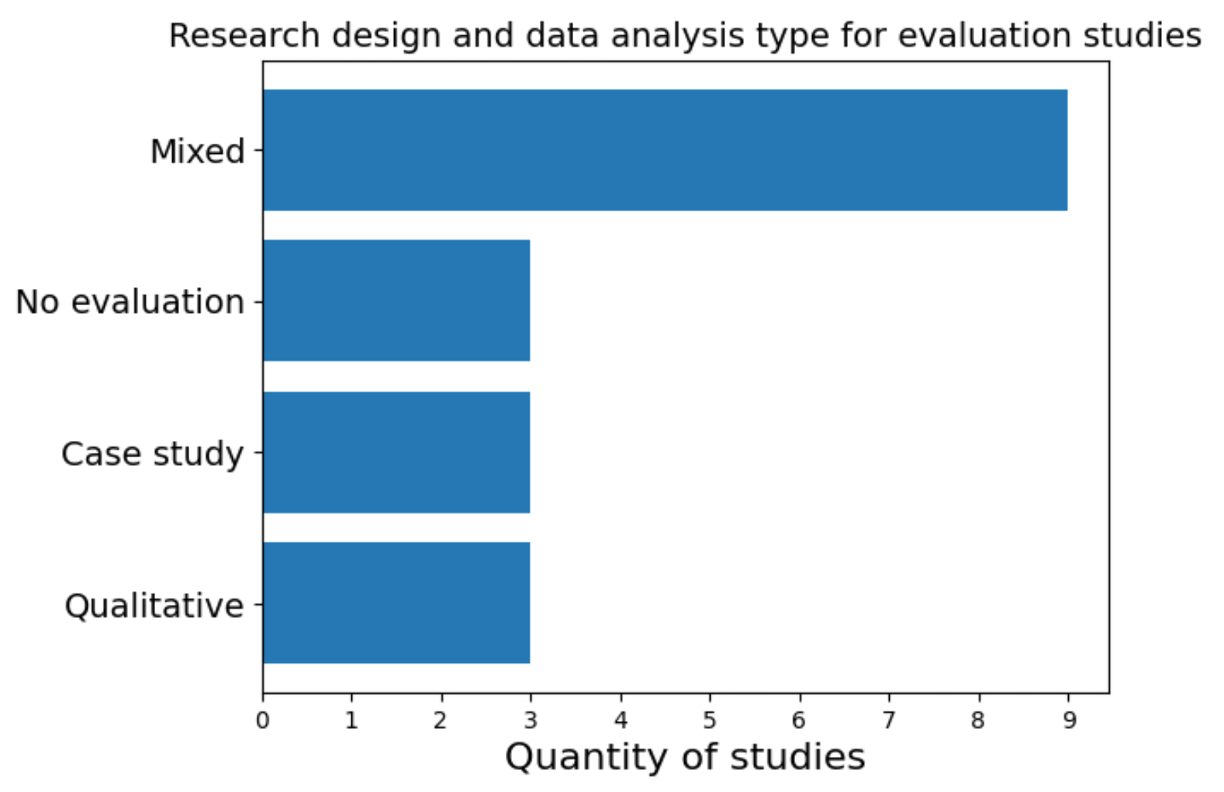}
    \caption{Research design and data analysis type}
    \label{fig:res_design}
  \end{minipage}
  \hfill
  \begin{minipage}[b]{0.45\textwidth}
    \includegraphics[width=0.85\textwidth]{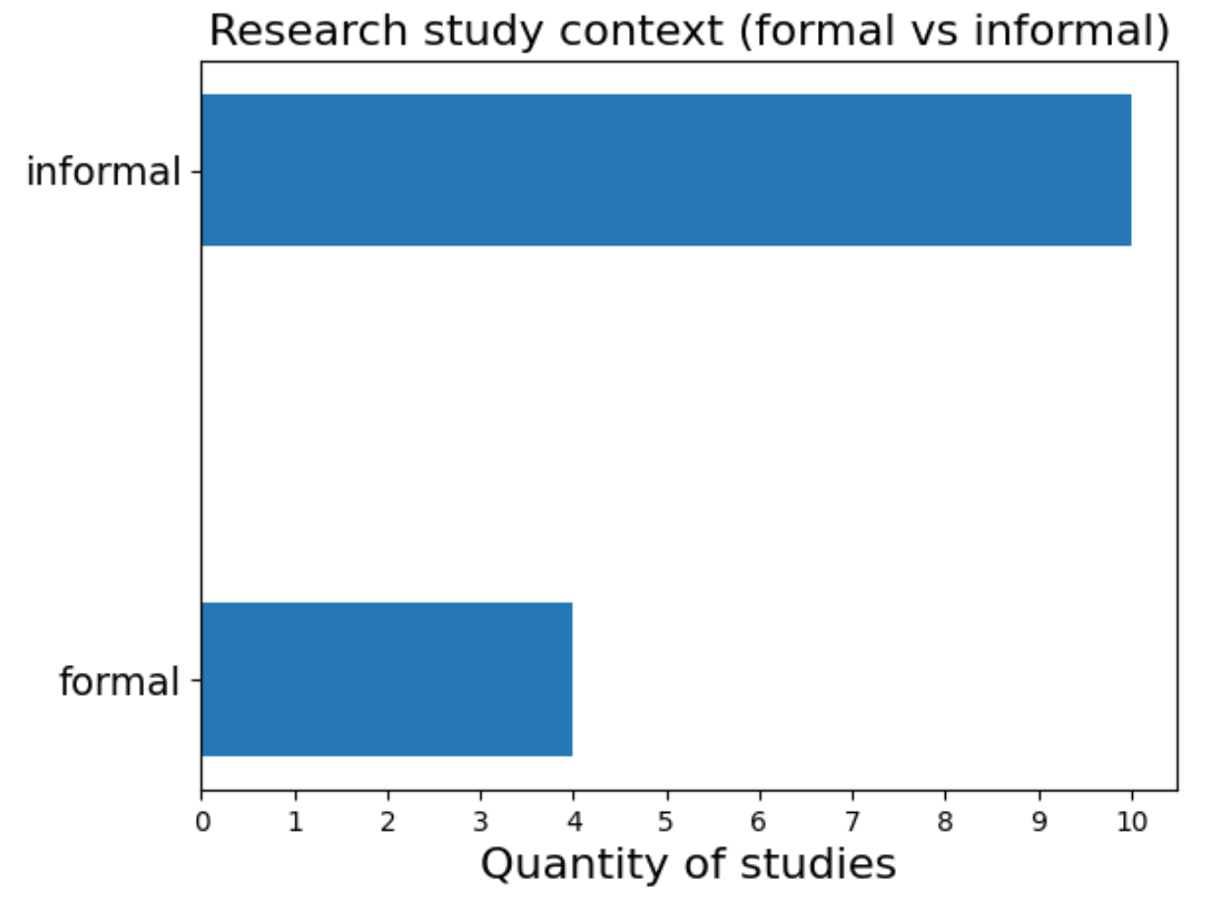}
    \caption{Research study context (formal vs. informal)}
    \label{fig:context}
  \end{minipage}

\end{figure}

Next, we report the research design and data analysis type for the learning tools that reported evaluations. As shown in Figure \ref{fig:res_design}, the majority (9/15) of studies reported their evaluation using a mixed methods approach. Other evaluations have been conducted as qualitative (3) or case studies (3). Sample sizes are moderate, ranging from 3 to 135 participants. The median sample size is 29.5 with a standard deviation of 36.9. Studies with relatively large sample sizes include \citet{rodriguez2021evaluation}, presenting a study with 135 students, \citet{druga2018growing}, presenting a multi-site international study including 104 children, as well as \citet{van2022learning}, with 49 participants including 27 children and 19 parents.

The research study contexts are categorized into informal (10) and formal learning environments (4), as shown in Figure \ref{fig:context}. Formal learning environment means study that is conducted in a classroom in a school, during the normal class time. In formal learning environments, the topics are integrated into relevant subjects (e.g., social studies). Students are expected to participate in the learning activity as they would in their normal class activities. In contrast, in informal learning environments such as after-school programs, online workshops and summer camps, students have much more freedom to choose whether to attend the activity or not. Because of the voluntary and low-pressure nature of the informal learning environments, the learning activities are typically designed with special consideration for keeping participants interested and engaged, especially for multiple-sessions or longitudinal studies \cite{druga2018growing}. 

The duration of the studies varied widely, with the shortest lasting only for 2 hours and the longest extending to 30 hours spanning over a month. The majority of the studies conducted their research over multiple sessions or days, often in the format of workshops or class meetings, with the duration of individual sessions ranging from 1.5 hours to 3.5 hours.

\subsubsection{Learning Tools Evaluation Outcomes}		

\begin{figure}[!htb]
  \includegraphics[width=0.8\textwidth]{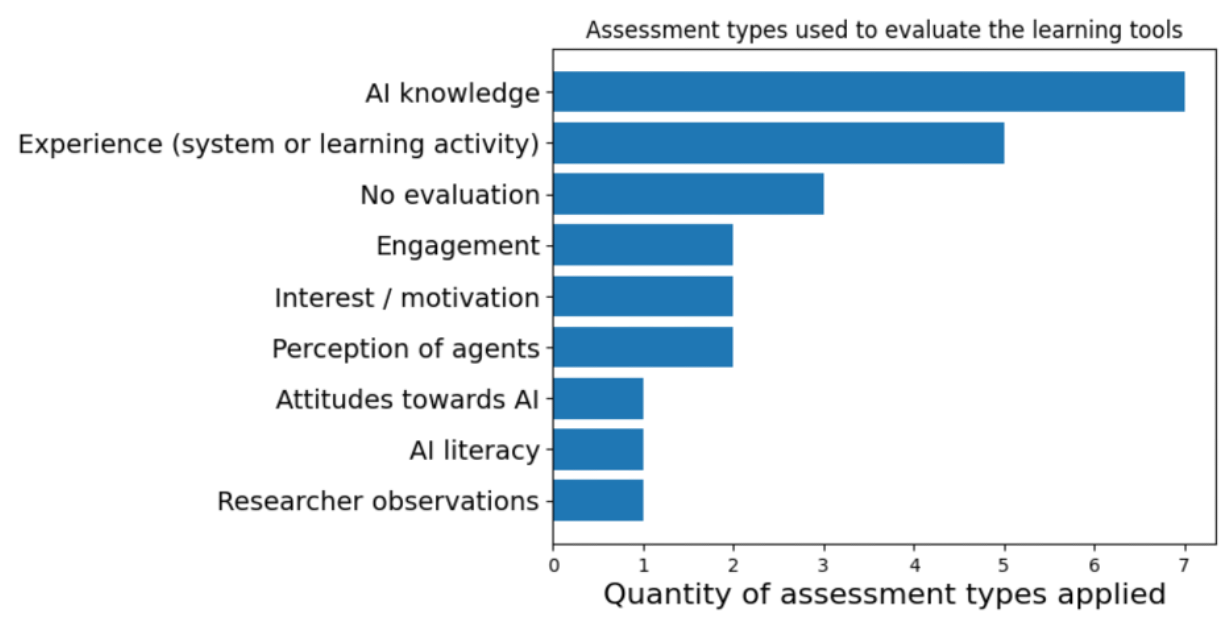}
  \caption{Assessment types used to evaluate the learning tools, one tool might use multiple assessment types}

  \label{fig:assess_type}
\end{figure}

The outcomes in evaluating learning tools primarily encompass the acquisition of AI knowledge, learning experiences, levels of engagement, and perceptions and attitudes towards AI (Figure \ref{fig:assess_type}). Among these, the most prevalent form of assessment is AI knowledge acquisition, as measured in seven tools. Next, we present the findings of these evaluative studies organized by the aforementioned outcomes.

\textbf{AI knowledge acquisition.} Studies consistently report a positive shift in learners’ AI knowledge. For instance, \citet{lin2020zhorai} evaluated children's understanding of the concepts introduced by Zhorai, concluding that a conversational interface coupled with visualization of the training process was effective in teaching children about machine learning concepts. In a similar vein, a study involving youth aged 10-16 years using LearningML \cite{rodriguez2021evaluation} reported significant improvement in participants' understanding of AI through comparison of pre- and post-tests (p < .001, effect size 0.486). Interestingly, this intervention's impact was particularly higher for learners with less prior AI knowledge (p < .001, effect size 1.007). In a study involving 40 students from Indonesia aged 16-17 using AI blocks in Snap! \cite{kahn2018ai}, the analysis of students' written essays found that 77.5\% of the essays demonstrated understanding of AI. From the essays, students also identified both benefits and dangers of speech synthesis. Similarly, in a series of user studies using Cognimates, \citet{druga2018growing} reported that children developed a strong understanding of AI concepts, but their ability to collaborate and communicate played a significant role in their learning, and their collaboration skills and understanding of AI concepts differed by the SES background of their community.

\textbf{Learning experience.} This phenomenon is another focal point of evaluation, five systems report the experience of system use and overall learning activity, assessed through system log analysis, reflective questionnaires and user feedback. This construct, inclusive of engagement as a measurable variable, revealed enriching experiences across diverse tools. For instance, the application of NLP4All in classrooms fostered meaningful participation and constructive discussions \cite{NaturalLanguageProcesing4All}, while the use of Teachable Machine (TM) in mobile environments demonstrated feasibility and accessibility for novice learners, albeit with limitations in predictive accuracy and sound recognition \cite{toivonen2020co}. The integration of physical devices like Amazon Echo was found engaging by \citet{van2021teaching}, highlighting their role in making AI accessible to everyone, even though they are not strictly necessary. A usability study involving Interactive Word Embeddings (IWE) pointed out that while 3D scatter plots and word analogy were intuitive and understandable, user-defined semantic dimensions required more explanations for proper interpretation \cite{bandyopadhyay2022interactive}. Another study using Teachable Machine (TM) in a primary school in Finland revealed the tool's feasibility for novice learners exploring machine learning, especially in mobile environments due to the low computational cost \cite{toivonen2020co}. However, limitations were noted in prediction accuracy due to small training data and in sound recognition models, which were partially impaired by background noise. Furthermore, the gamification and competitive elements integrated into AI tools can significantly improve students' understanding and engagement with AI, contributing to a richer and more immersive learning experience \cite{NaturalLanguageProcesing4All}.

\textbf{Perceptions, attitudes, and interests.} The cultivation of interest and the shaping of perceptions and attitudes towards AI are crucial outcomes explored in several studies. These studies highlight the importance of designing agent persona and providing transparency and explanation to build trust in conversational agents \cite{van2022learning}. The findings regarding perceptions of AI are mixed. In a study in which students learned about and utilized Convo, their perceptions of the intelligence of conversational AI agents decreased post-workshop, while their confidence in their abilities to construct these agents increased \cite{zhu2021teaching}. In contrast, \citet{van2021alexa} report students
felt Alexa to be more intelligent and felt closer to Alexa after programming the conversational AI. Another study raised concerns about over-trusting technology. \citet{van2022learning} observe that children with no programming experience reported higher trust in Alexa’s correctness after a programming activity.

\section{Case study: ConvoBlocks} \label{sec:case-study}

In this section, we describe a well-studied system, ConvoBlocks on MIT App inventor, as a concrete example of how a specific NLP teaching tool is developed and evaluated. 
 
\subsection{Overview of ConvoBlocks}
ConvoBlocks is a block-based programming environment for 5-12th grade students to create conversational agents. ConvoBlocks is primarily designed for students aged 11 to 18 to create their own conversational agents for deployment on any Alexa-enabled device or within the MIT App Inventor interface. The programming interface includes a range of code block types: ``Voice,'' which triggers Alexa to execute a task such as speaking a phrase; ``Control,'' which modifies the program's flow based on constructs such as ``if'', ``while''; and ``Text,'' which allows textual data manipulation, such as combining strings.

\subsubsection{Pedagogical Approach} 
\begin{figure}[!htb]
  \includegraphics[width=0.9\textwidth]{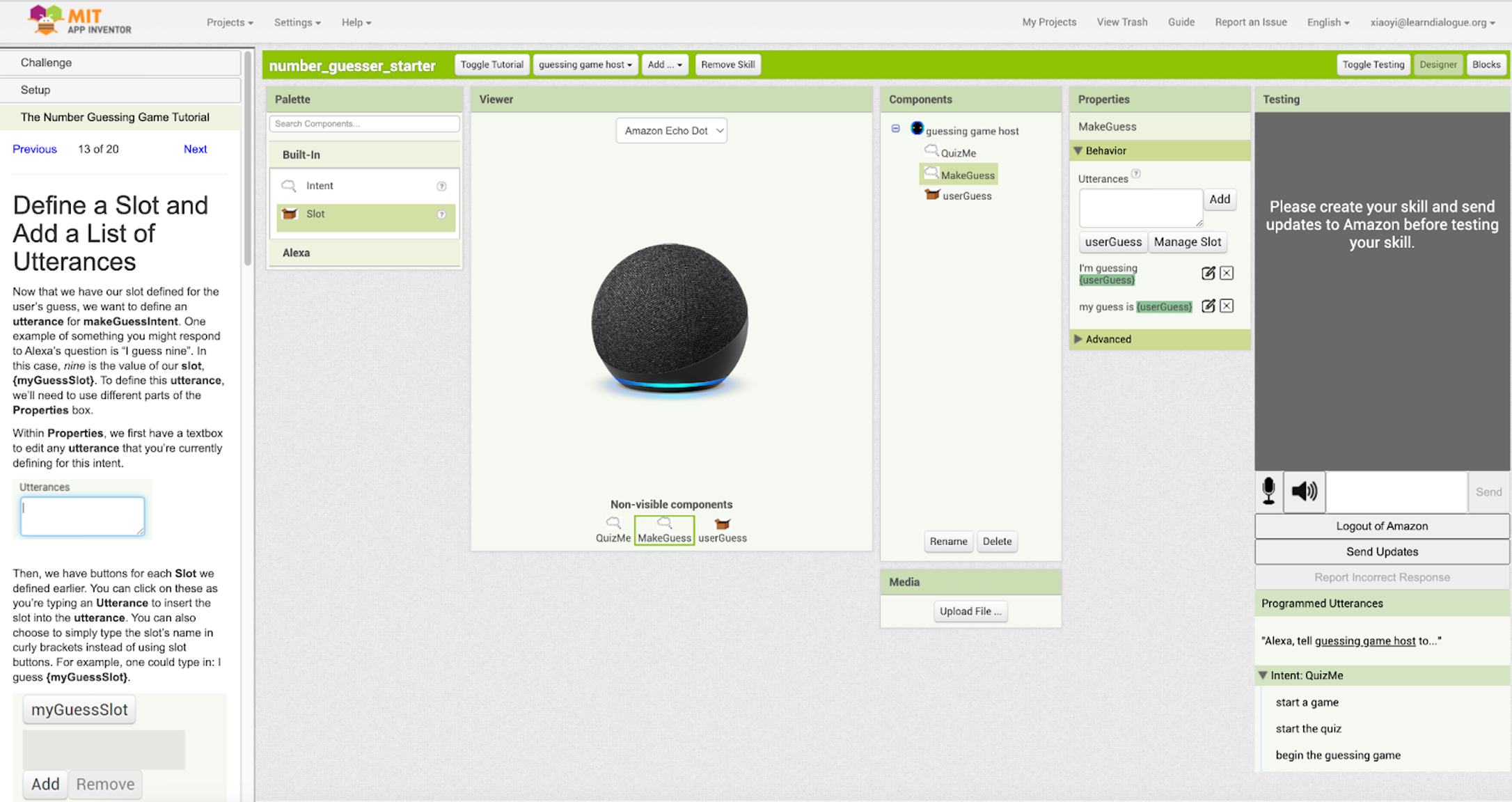}
  \caption{ConvoBlocks interface, ``Designer'' view}

  \label{fig:case_1}
\end{figure}

\begin{figure}[!htb]
  \includegraphics[width=0.9\textwidth]{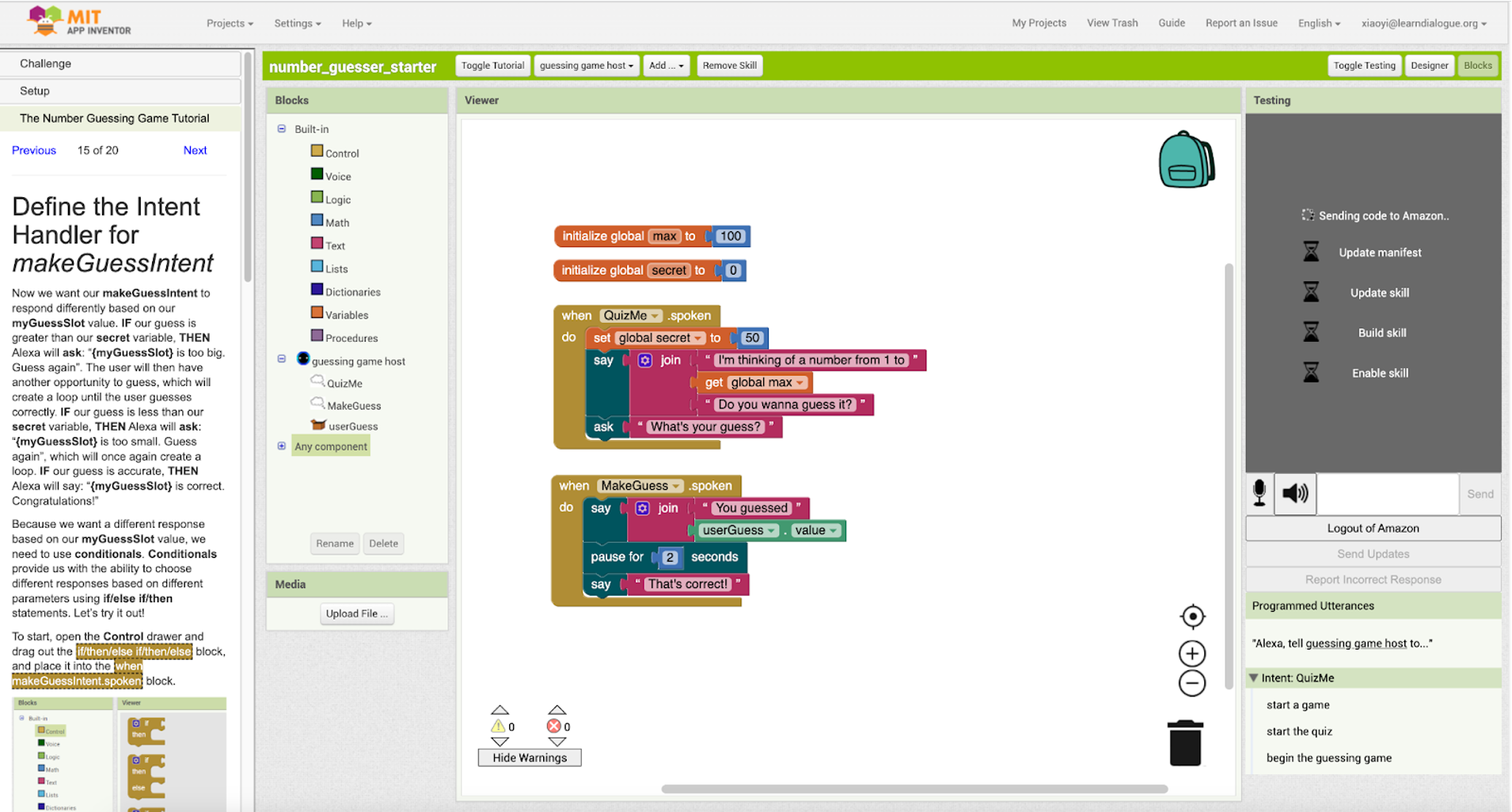}
  \caption{ConvoBlocks interface, ``Blocks'' page}

  \label{fig:case_2}
\end{figure}

ConvoBlocks offers a list of built-in tutorials for beginner, intermediate and advanced learners. For beginners, the system provides ``Alexa Hello World'' and ``Alexa Calculator'' tutorials. The system provides an ``Alexa number guessing game'' tutorial for intermediate learners and an ``Alexa Messenger'' tutorial for advanced learners. Figure \ref{fig:case_1} and \ref{fig:case_2} show the interface for the intermediate ``number guessing game'' tutorial, in which the target grade level is 6-12th grade. This tutorial guides the developer to create an Alexa Skill that allows the player to try guessing a random number while receiving feedback on whether the guessed number is too big, too small or just right. Within the development environment, the tutorial is positioned at the left, and the user has the option to toggle the display of the tutorial by clicking the ``Toggle Tutorial'' button. The tutorial is divided into several tabs, and each tab represents a module (e.g., setup, programming, integration), toggled vertically. There are multiple pages in each tab, toggled horizontally; each contains a fine-grained sequence of actions and screenshots to complete the task. 

\subsubsection{System Interface Design}
For the project working page, the programming environment includes two views, a ``Designer’’ view (Figure \ref{fig:case_1}) and a ``Blocks’’ page (Figure \ref{fig:case_2}). In the Designer view, developers set high-level project types, function names, intents and entities; in the Blocks page, developers program the specific actions the artifact will perform (e.g., say ``Hello world'') by dragging and dropping block-based code elements. The user can test the project on the right side of the screen in both views.

\subsection{Evaluation and Critical Analysis}
\subsubsection{Study Implementation and Outcomes}
Besides the system's built-in tutorial and its own block-based programming interface, the research team of ConvoBlocks also developed a conversational AI curriculum that was implemented as a 5-day, 2.5-hour-per-day workshop. On a high level, the workshop covers important topics in conversational AI and NLP, including training, transfer learning, large language models, intents, speech recognition and speech synthesis. The workshop schedule features a series of interactive lectures and group discussions on the ``Big 5 AI Ideas \citet{touretzky2019envisioning},'' comparing differences between rule-based and ML-based architectures, building conversational AI projects, and AI ethics. 

The main conversational agent concepts taught in the ConvoBlocks curriculum include events (dialogue management), conditions (also part of dialogue management), and event-driven program representations (conversation representation). Students define events and responses to manage agent dialogue, add conditions to their event handlers to define whether or not particular pieces of dialogue or events occur, and represent conversations through event-driven block-based programming. 

The ConvoBlocks environment was implemented and evaluated in informal learning environments. Preliminary results were published in 2021 \cite{van2021teaching}, in which the authors organized a five-day workshop with 47 students ranging in age from 11 to 18. The workshop achieved the desired outcome: significant learning gains were reported in general AI and conversational AI concepts, though machine learning and ethics were identified as more challenging areas. A total of 70 projects were created by the students. Programming conversational AI also seemed to make a difference in students’ perception of AI. Another evaluation study reported that after programming the conversational AI, students felt Alexa to be more intelligent and felt closer to Alexa \cite{van2021alexa}. This finding contrasts with \cite{zhu2021creating}'s study where they found participants' perceptions of the intelligence of conversational AI agents decreased after the programming workshop.

\subsubsection{Comparative Analysis}
While Convo allows for easy device integration and features a variety of scaffolding integrated in the environment, there are several disparities when compared to other systems in the landscape of NLP learning tools. To elaborate, one notable absence in ConvoBlocks is the ability to define conditional relationships between different intents, which is a feature often included in other conversational AI platforms such as Dialogflow \citet{dialogflow_cx}. In human-human conversations, certain dialogues are often contextually dependent on preceding interactions. For instance, a user might express the intent of \textit{request movie recommendation}, followed by a \textit{request more information} intent. However, ConvoBlocks treats all intents with uniform priority, not allowing for such contextual dependencies and consequently, lacks conversational flow--a visual representation that is especially pivotal in K-12 contexts \cite{park2013modeling, large2005interface}. This may make it challenging to create coherent and engaging conversational experiences, particularly for beginners. Balancing visual appeal with functional robustness is crucial, and understanding these disparities can guide future developments in creating more refined and inclusive learning tools.

% This comparative analysis aims not to diminish ConvoBlocks but to highlight the diverse range of features in NLP learning tools. 

\section{Discussion} \label{sec:discussion}

In this section, we will first summarize the key findings from the literature review and the case study. Then, we will identify gaps in the current state of research and design and discuss implications for NLP education in K-12.

\subsection{Key Findings}
This paper identified 11 digital learning environments for NLP learning in K-12 education, with most being accessible as online web apps. However, some tools have limitations such as restricted access or language support, which may affect their usability for beginners.

The 11 digital learning environments for NLP in K-12 education primarily support text classification, speech recognition, and intent recognition tasks, with limited support for other popular NLP tasks. These tools offer varying capabilities for training and deploying NLP models and provide different data input modalities, such as keyboard and speech. However, the majority of the systems lack in-depth scaffolding and explanations for NLP processes, which could be improved for better learner understanding.

A majority of the studies employed mixed methods for evaluating their tools, with moderate sample sizes ranging from 3 to 135 (median = 29.5). Research studies were more often deployed in informal learning contexts than formal contexts. 

Most NLP learning activities target middle and high school students, with evaluations focusing on AI knowledge assessment and learning experiences. These tools prove effective for teaching NLP and AI concepts, fostering interest, and improving students' understanding and engagement. However, learning challenges persist in machine learning and ethics concepts, and there is more work to be done in addressing issues of over-trusting technology. 

\subsection{Gaps and Implications}
In light of these findings, we identified several research and design gaps that need to be addressed:

\subsubsection{Gap 1: Limited NLP task variety}

Most of the NLP tasks supported by these tools are natural language understanding (NLU) related (e.g., intent recognition), with other NLP tasks such as speech synthesis, semantic parsing, and question-answering only introduced by a few tools. Given the breadth of NLP tasks, researchers and educators have to make choices about where to begin to teach children. NLU is a popular starting point because its applications (e.g., voice assistants) are ubiquitous in students' daily life, and thus can serve as a more tangible and relatable introduction to NLP. However, this review revealed a potential research gap in exploring learning tools to support NLP tasks other than NLU. While in-depth mastery of each NLP task is not the primary goal of K-12 NLP education, students should gain a foundational understanding of the range of the problems NLP can address, and be able to explore the more complex applications if they wish. There is a clear opportunity to develop and investigate new tools that support other important NLP tasks, such as text generation, information retrieval, and topic modeling, in order to expose younger students to a wider range of NLP tasks.

\subsubsection{Gap 2: Challenges in comprehensive evaluation methods}

Developing effective evaluation for NLP learning tools, especially for younger students, is fraught with challenges. Because many of the NLP learning tools are emergent and still being refined, they have not been extensively evaluated. Among the tools that have received evaluation, this evaluation predominantly leans towards AI knowledge assessment, with specific emphasis or qualitative reports on student learning experiences. The learning objectives of the activities aim to foster students' general AI knowledge. As a result, there is limited or no emphasis on evaluating students’ understanding of NLP concepts in the assessments. 

Evaluating educational tools in general, particularly in K-12 environments, presents many challenges \cite{falloon2020understanding}. One of the primary obstacles is student recruitment. Gaining access to, and permissions for, student participation can be a logistical challenge. Other considerations range from aligning with school calendars, managing sample sizes, and deciding on the format and duration of the program. Integrating these learning tools into the existing school ecosystem also poses dilemmas: How does one best incorporate them? Is it more effective to introduce them as voluntary add-ons, or replace current curriculum components? Should the evaluation prioritize consistent administration by having dedicated researchers present, or should it prioritize authenticity and work toward scalability by training teachers to manage these new educational modules themselves?

Given these challenges, it is unsurprising that a robust, NLP-specific evaluation mechanism is not  commonplace yet. As the field progresses, refining evaluation methods to holistically assess NLP-specific learning outcomes will be crucial. This will not only ensure the effectiveness of the learning tools but also enhance our understanding of their impact on students' learning trajectories.

\subsubsection{Gap 3: Insufficient pedagogical explanation}

This review reveals that while many NLP and ML pedagogical systems provide some level of explanation to support learner understanding of tasks and processes, there is still room for improvement. Model accuracy is the most popular evaluation metric presented to users, but not all systems offer detailed evaluation outputs or interactive explanations. The question of how to distill an abstract evaluation metric into intuitive insights, without oversimplifying the underlying mechanism, remains a challenge. Researchers might draw on work in non-NLP applications, where using explainability techniques and interpretable ML models has been shown to enhance children's understanding of AI \cite{melsion2021using}. To better support users in developing a comprehensive understanding of NLP and ML techniques, pedagogical systems should refine their evaluation metrics, adopt interpretable models, offer more in-depth explanations of model workings, and incorporate engaging educational methods, such as animations and visual aids \cite{alm2021visualizing,kahng2018gan}.

\subsubsection{Gap 4: Limited focus on younger children}

The majority of NLP and ML learning activities target middle and high school students, with only a few studies focusing on children under the age of 11. Given the complexity of NLP, which requires knowledge in programming, linguistics, and mathematics, it seems reasonable to introduce NLP concepts to students with a stronger foundation in these areas. Besides the complexity of the content, gaining consistent access to younger age groups can be challenging due to protective educational environments and parental concerns. Additionally, these children exhibit more considerable variation in developmental readiness, which further challenges the design and evaluation of suitable learning activities. However, this should not exclude younger students from learning about NLP. There is a gap in learning activities specifically designed for younger children. In future studies, researchers and educators should develop tools and activities adaptable to younger age groups and their level of understanding in order to cultivate early interest in NLP and ML.

\subsubsection{Gap 5: Insufficient personalized support for diverse learners}

The current learning tools rarely deliver personalized and adaptive learning experiences to cater to the unique needs of individual learners. Given the significant variation in cognitive load and learning requirements between beginner and advanced activities \cite{bounajim2021applying}, a one-size-fits-all approach can lead to less effective learning outcomes. This marks a critical gap in the field, underscoring the need for tools that offer adaptivity, accommodating learners' skill levels through personalized levels of difficulty, scaffolding, and explanation \cite{li2021features}. Future research should prioritize the development of systems that can dynamically adjust to individual learners, thereby fostering a more inclusive and effective learning environment. 

\subsubsection{Gap 6: Lack of recommendations for effective teaching strategies}

While some studies offer insights into successful teaching methods, there is a gap in the literature regarding concrete recommendations for effectively incorporating NLP education in K-12 classrooms. Strategies such as gamification, embodiment, iteration, and instant feedback have been shown to enhance learning experiences \cite{baylor2003effects,gresse2021visual}, but these need to be further explored and integrated into NLP educational tools. A complementary strategy, using relevant data that relates to students' lives or real-world problems, can create more engaging and meaningful learning experiences \cite{garcia2020learningml,register2020learning}. For example, training speech recognition systems with user-defined speech data can empower students to develop socially relevant agents that are sensitive to gender, voice, and accent variations. By providing educators and developers with clear recommendations for effective teaching strategies, NLP learning tools can be designed to better engage and educate students in this complex field.

\section{Conclusion}

NLP education in K-12 presents a unique opportunity to engage students in the interdisciplinary field of language and technology. In this paper, we present a comprehensive literature review of the state-of-the-art learning environments that support teaching of NLP concepts. We summarized key findings from the development of these learning tools and their evaluations. We also highlight the current gaps in the field of NLP learning tools and activities, and we discuss practical implications and future directions. Using these design implications, we hope that educators and researchers can develop more effective and engaging educational experiences for a broader range of students. Diversifying NLP tasks, refining evaluation methods, enhancing explanations and scaffolding, and creating age-appropriate learning activities will not only improve students' understanding of NLP concepts but also foster early interest in this rapidly evolving field. As technology continues to advance, it is crucial to equip the next generation with the knowledge and skills necessary to navigate and contribute to the world of NLP and AI. 

\begin{acks}

I would like to thank Amogh Mannekote and Lydia Pezzullo for their thorough copy-editing of this paper, and to everyone in the LearnDialogue group for their companionship and encouragement.
\end{acks}

\bibliographystyle{ACM-Reference-Format}
\bibliography{references}

%%
%% If your work has an appendix, this is the place to put it.
% \appendix

% \section{Summary of evaluation of the tools}

\end{document}